\documentclass[11pt]{article}

\usepackage{acl}

\usepackage{times}
\usepackage{latexsym}
\usepackage[T1]{fontenc}
\usepackage[utf8]{inputenc}
\usepackage{microtype}
\usepackage{inconsolata}
\usepackage{graphicx}
\usepackage{booktabs}
\usepackage{multirow}
\usepackage{amsmath}
\usepackage{amssymb}
\usepackage[table]{xcolor}
\usepackage{subcaption}
\usepackage{enumitem}
\usepackage{CJKutf8}
\newcommand{\ko}[1]{\begin{CJK}{UTF8}{mj}#1\end{CJK}}

\title{KVoiceBench, KOpenAudioBench, and KMMAU:\\Agent-Driven Korean Speech Benchmarks for Evaluating SpeechLMs}

\author{
{\bfseries Haechan Kim$^{1,2}$ \quad
Seungjun Chung$^{1}$ \quad
Inkyu Park$^{1}$} \quad
Jihoo Lee$^{1,3}$ \quad
Jonghyun Lee$^{1}$\thanks{\;Corresponding author.} \\
\\[-0.2em]
$^{1}$KRAFTON \\
$^{2}$Kim Jaechul Graduate School of AI, KAIST \\
$^{3}$Department of Mathematical Sciences, Seoul National University \\
\\[-0.5em]
\texttt{\{kim.haechan2,s.j.chung,inkyupark,numbering,jonghyunlee\}@krafton.com}\\[4.0em]
}

\begin{document}
\maketitle

\begin{abstract}
Speech language models (SpeechLMs) have achieved substantial progress by extending large language models (LLMs) to the speech modality. However, SpeechLM evaluation remains heavily centered on English, limiting reliable assessment of multilingual speech capabilities. Straightforward benchmark transfer through ASR, translation, normalization, and TTS can corrupt language-specific instructions, answer constraints, and spoken forms; for audio understanding, transferring source-language audio also fails to preserve target-language speaker attributes, accents, and paralinguistic properties. To address these limitations, we propose two human-agent benchmark-construction frameworks: one transfers source-language SpokenQA benchmarks into target-language SpokenQA benchmarks, and the other converts target-language ASR corpora into audio understanding benchmarks using transcriptions and speaker metadata. Using these frameworks, we construct and publicly release three Korean speech benchmarks: KVoiceBench\footnote{\url{https://huggingface.co/datasets/KRAFTON/KVoiceBench}} and KOpenAudioBench\footnote{\url{https://huggingface.co/datasets/KRAFTON/KOpenAudioBench}} for Korean SpokenQA, and KMMAU\footnote{\url{https://huggingface.co/datasets/KRAFTON/KMMAU}} for Korean audio understanding, comprising 12,345 samples in total. We evaluate eight recent SpeechLMs and find that English–Korean performance gaps vary substantially across models and task families, and that SpokenQA and audio understanding rankings diverge, revealing complementary weaknesses invisible to English-only evaluation\footnote{Our evaluation code is available at: \url{https://github.com/krafton-ai/Raon-Eval}}.
\end{abstract}

\section{Introduction}

Recent advances in large language models (LLMs) have accelerated the development of speech language models (SpeechLMs), which extend LLM capabilities to spoken and audio interaction through speech encoders, audio tokenizers, and speech generation modules \citep{rubenstein2023audiopalm,zhang2023speechgpt,tang2024salmonn,chu2023qwenaudio,xu2025qwen25omni,kimiTeam2025kimiaudio}. As these models move from transcription-oriented systems toward voice assistants and audio-interactive agents, evaluation must test not only automatic speech recognition but also reasoning, instruction following, safety, and grounded understanding of audio inputs \citep{chen2024voicebench,li2025baichuanaudio,yang2024airbench,wang2025audiobench}.

Spoken question answering (SpokenQA) and audio understanding have therefore become central evaluation settings for SpeechLMs. SpokenQA evaluates whether a model can answer questions delivered in speech, while audio understanding evaluates semantic and paralinguistic information contained in the audio signal, including speaker attributes, emotion, acoustic scenes, and music \citep{li2018spokensquad,wu2024heysquad,ao2024sdeval,sakshi2024mmau,wang2025audiobench}. However, most widely used SpeechLM benchmarks remain heavily centered on English. Multilingual resources such as SD-QA and FLEURS improve coverage for dialectal spoken QA, ASR, language identification, translation, and retrieval, but they do not provide a scalable framework for transferring modern SpokenQA benchmarks or constructing target-language audio understanding benchmarks \citep{faisal2021sdqa,conneau2023fleurs}.

A common strategy for expanding text benchmarks is to translate English data into target languages, often with professional translation, machine translation, post-editing, or LLM-based translation \citep{conneau2018xnli,lewis2020mlqa,ponti2020xcopa,ahuja2023mega,xuan2025mmluprox}. A direct speech analogue is a cascade that translates source-language transcripts, normalizes the translated text, and synthesizes target-language speech with TTS, resembling multilingual speech-translation resources and speech-to-speech corpora that pair translation with synthesized speech \citep{wang2020covost2,jia2022cvss}. While simple and scalable, such pipelines inherit known translation artifacts and can fail to preserve task-relevant linguistic properties \citep{artetxe2020translation,clark2020tydiqa}. For example, the English instruction ``Write all letters in upper case'' cannot be meaningfully transferred to languages such as Korean that do not encode uppercase--lowercase distinctions.

Speech synthesis also introduces a second source of benchmark invalidity: text normalization. TTS systems require written text to be converted into speech-friendly forms, but numbers, dates, abbreviations, and context-dependent readings are long-standing hard cases for both rule-based and neural normalization systems \citep{sproat2001normalization,ebden2014kestrel,zhang2019neuraltextnorm}. These issues are especially harmful for SpokenQA, where a small normalization error can change the answerability of the question itself. Audio understanding adds a different constraint: speaker identity, accent, emotion, overlap, and other paralinguistic properties are properties of the waveform rather than properties of a translated transcript \citep{faisal2021sdqa,ao2024sdeval,wang2025audiobench}.

To address these limitations, we propose two human-agent collaborative frameworks for constructing high-quality target-language speech benchmarks and instantiate them as a Korean benchmark suite. The first framework converts source-language SpokenQA benchmarks into target-language SpokenQA benchmarks through four stages: ground-truth correction, hypertranslation, speech-friendly normalization, and TTS synthesis (Figure~\ref{fig:kvoicebench_pipeline}). In the ground-truth correction stage, two frontier LLM agents operating as a reviewer and a meta-reviewer identify and correct erroneous annotations in existing source-language benchmarks. The core component, hypertranslation, uses a rulebook that explicitly encodes grammatical, orthographic, and writing-system-specific properties of the target language. This rulebook is iteratively constructed through a human-agent collaborative loop, enabling systematic handling of language-specific edge cases. The normalization stage converts hypertranslated text into speech-friendly forms suitable for TTS synthesis, and the normalized text is then synthesized into speech.

The second framework constructs target-language audio understanding benchmarks from naturally occurring target-language ASR corpora rather than transferring source-language audio (Figure~\ref{fig:kmmau_pipeline}). It leverages audio, transcriptions, and speaker metadata, and selects the construction method according to the capability being tested: rule-based generation from speaker metadata for acoustic attributes, rule-based generation from transcriptions for lexical questions, LLM-generated questions with human review for semantic understanding, and fully manual annotation for holistic capabilities that require listening to the audio. This design enables both semantic and paralinguistic question-answer pairs to be grounded in authentic target-language speech.

\begin{figure*}[t]
\centering
\includegraphics[width=\textwidth]{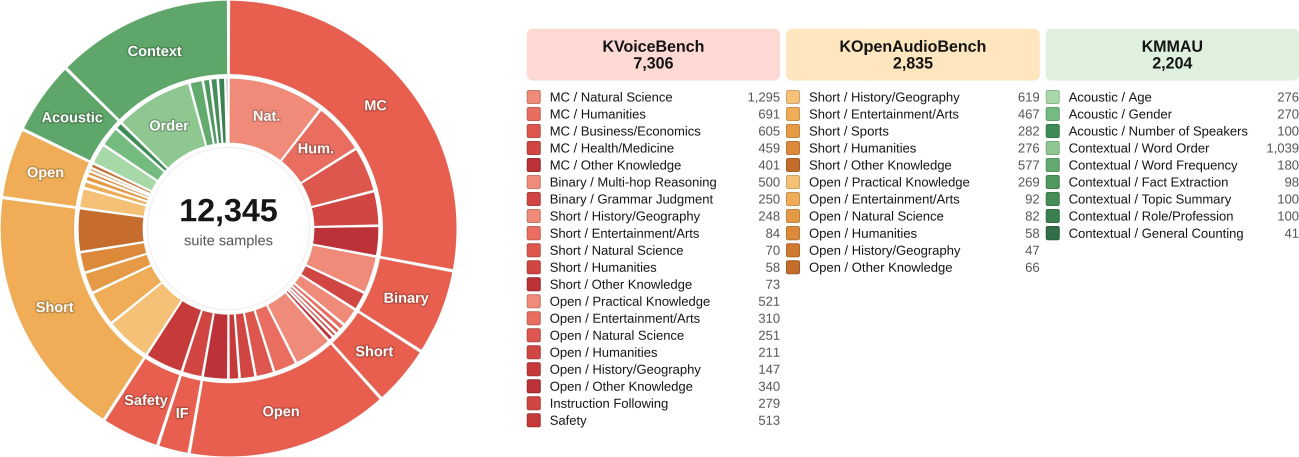}
\caption{Task-family and fine-grained category distribution of the released Korean speech benchmark suite. The outer ring groups samples into task families, the inner ring shows detailed categories and capabilities, and the legend reports sample counts for KVoiceBench, KOpenAudioBench, and KMMAU.}
\label{fig:benchmark_distribution}
\end{figure*}

Using these frameworks, we construct and publicly release three Korean speech benchmarks (Figure~\ref{fig:benchmark_distribution}): two SpokenQA benchmarks, KVoiceBench and KOpenAudioBench, and one audio understanding benchmark, KMMAU. KVoiceBench and KOpenAudioBench are derived from the English SpokenQA benchmarks VoiceBench and OpenAudioBench, respectively, and contain 7,306 and 2,835 samples \citep{chen2024voicebench,li2025baichuanaudio}. During the two SpokenQA transfers, 578 of 10,719 source samples are rejected during curation (5.4\%). KMMAU is constructed from Korean ASR corpora including KSS, KMSAV, and Seoul Corpus, and consists of 2,204 samples \citep{park2018kss,park2024kmsav,yun2015seoul}. These benchmarks support more reliable multilingual SpeechLM evaluation.

We also evaluate eight SpeechLMs to analyze how current models behave across English and Korean SpokenQA and audio understanding. The results show that Korean SpokenQA performance drops substantially relative to English, but the degradation is not uniform across models or task families. Audio understanding shows a different ranking pattern from SpokenQA, suggesting that target-language question answering and naturally grounded target-language audio understanding probe complementary capabilities.

Our contributions are as follows:

\begin{enumerate}
    \item We propose reproducible human-agent collaborative frameworks for transferring SpokenQA benchmarks and constructing target-language audio understanding benchmarks.
    \item We construct and publicly release three Korean speech benchmarks---KVoiceBench, KOpenAudioBench, and KMMAU---with rulebooks that make the construction process auditable and reusable by native speakers of other target languages.
    \item We evaluate eight recent SpeechLMs and find that English–Korean performance gaps vary substantially across models and task families, and that SpokenQA and audio understanding rankings diverge, revealing complementary weaknesses invisible to English-only evaluation.
\end{enumerate}


\section{Related Work}

\paragraph{Speech Language Models.}
SpeechLMs connect LLMs with speech and audio representations so that models can process spoken input and, in some systems, generate spoken responses. Early systems such as SpeechGPT and AudioPaLM explored speech-text interaction by combining LLM reasoning with speech units or speech-to-text/text-to-speech components \citep{zhang2023speechgpt,rubenstein2023audiopalm}. Later audio-language models such as SALMONN and Qwen-Audio broadened the input space from speech to general audio, including environmental sounds and music \citep{tang2024salmonn,chu2023qwenaudio}. Recent omni-modal and audio foundation models further integrate audio understanding, speech generation, and streaming interaction \citep{xu2025qwen25omni,kimiTeam2025kimiaudio,li2025baichuanaudio}. This shift motivates evaluation protocols that go beyond ASR accuracy and test whether the model can reason over what is said and how it is spoken.

\paragraph{SpokenQA and Audio Understanding Benchmarks.}
SpokenQA extends text question answering to spoken inputs. Spoken SQuAD and HeySQuAD showed that ASR errors and human-spoken questions substantially affect downstream QA performance \citep{li2018spokensquad,wu2024heysquad}, and SD-QA introduced a multi-dialect spoken QA benchmark covering five languages and 24 dialects \citep{faisal2021sdqa}. More recent benchmarks such as VoiceBench and OpenAudioBench evaluate LLM-based voice assistants on reasoning, knowledge, instruction following, safety, and open-ended questions \citep{chen2024voicebench,li2025baichuanaudio}. In parallel, audio understanding benchmarks such as AIR-Bench, SD-Eval, MMAU, MMAU-Pro, and AudioBench evaluate interaction with broader audio signals, including speech, sound events, speaker properties, and music \citep{yang2024airbench,ao2024sdeval,sakshi2024mmau,kumar2025mmaupro,wang2025audiobench}. Our work targets the intersection: transferring SpokenQA benchmarks and converting ASR data into audio understanding benchmarks for a target language.

\paragraph{Multilingual Benchmark Construction.}
Multilingual text evaluation often relies on translating English benchmark items into other languages, as in XNLI, MLQA, XCOPA, MEGA, and MMLU-ProX \citep{conneau2018xnli,lewis2020mlqa,ponti2020xcopa,ahuja2023mega,xuan2025mmluprox}. Other benchmarks, such as TyDi QA, emphasize native data collection to better capture typologically diverse language phenomena that may be absent from English-centered datasets \citep{clark2020tydiqa}. Prior work has also shown that translated benchmark data can contain artifacts that alter model behavior and weaken cross-lingual conclusions \citep{artetxe2020translation}. For speech, resources such as FLEURS, CoVoST 2, and CVSS provide multilingual speech data for ASR, language identification, speech translation, and speech-to-speech translation \citep{conneau2023fleurs,wang2020covost2,jia2022cvss}. They do not, however, solve the problem of transferring SpeechLM evaluation samples whose validity depends on language-specific instructions, orthography, and paralinguistic cues. Our hypertranslation framework is designed for this benchmark-validity problem rather than for ordinary sentence-level translation.

\paragraph{Text Normalization for Speech Synthesis.}
Text normalization is a core preprocessing step for TTS: written tokens such as dates, numbers, abbreviations, measurement expressions, and symbols must be verbalized in contextually appropriate spoken forms \citep{sproat2001normalization,ebden2014kestrel}. Both neural and rule-based systems can achieve high aggregate accuracy while still making errors that are unacceptable for speech applications, especially when rare constructions or context-sensitive readings determine the intended meaning \citep{sproat2017rnntextnorm,zhang2019neuraltextnorm}. In benchmark construction, such errors can change the answer or make a spoken question unanswerable, so we treat normalization as a dedicated stage guided by target-language rules.

\section{Benchmark Construction Framework}
\label{sec:framework}

\subsection{SpokenQA Benchmark Transfer}
\label{sec:spokenqa_framework}

\begin{figure*}[t]
\centering
\includegraphics[width=\textwidth]{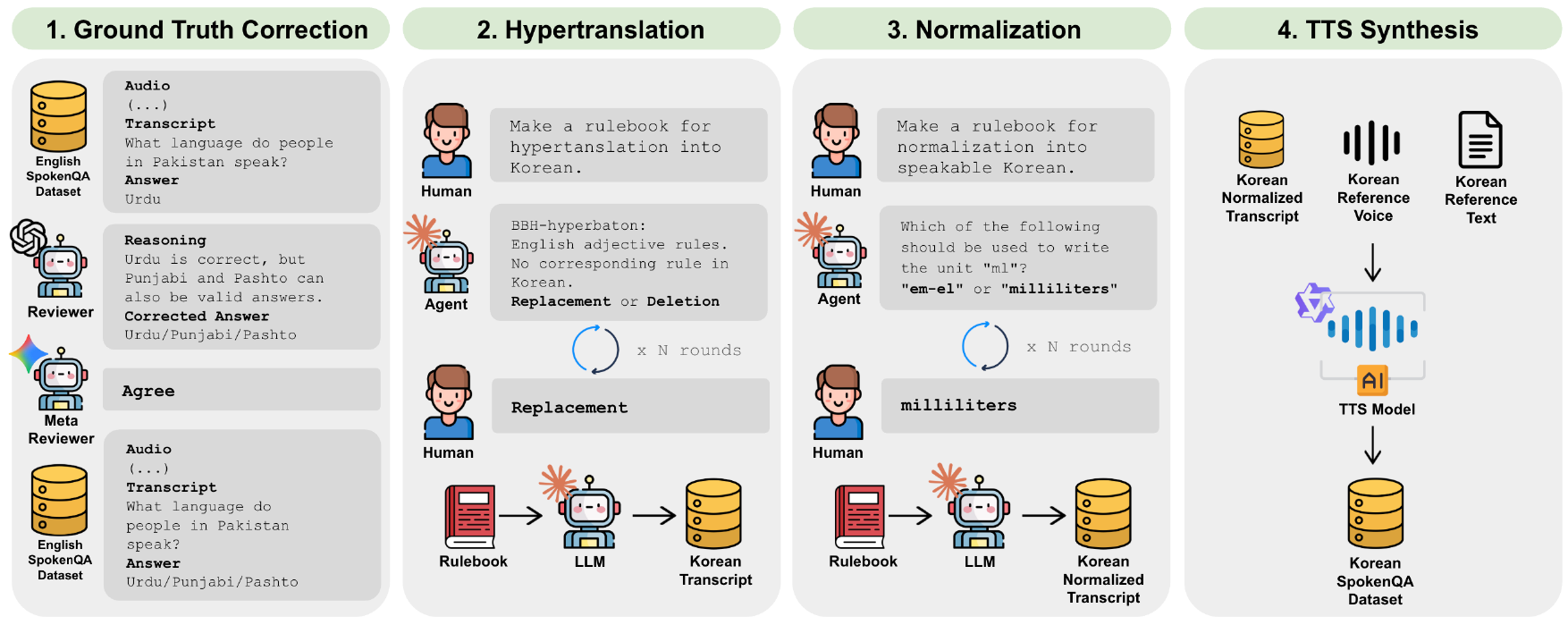}
\caption{Construction pipeline for KVoiceBench and KOpenAudioBench. The framework consists of four stages: (1) ground-truth correction via reviewer and meta-reviewer LLMs, (2) hypertranslation guided by a human-agent rulebook, (3) speech-friendly normalization guided by a separate normalization rulebook, and (4) TTS synthesis using Korean reference voices.}
\label{fig:kvoicebench_pipeline}
\end{figure*}

This framework transfers source-language SpokenQA benchmarks into a target language.
As shown in Figure~\ref{fig:kvoicebench_pipeline}, the framework consists of four stages: ground-truth correction (\S\ref{sec:gt_correction}), hypertranslation (\S\ref{sec:translation}), speech-friendly normalization (\S\ref{sec:normalization}), and TTS synthesis (\S\ref{sec:tts}).
In this work, we apply this pipeline to English-to-Korean transfer, converting VoiceBench and OpenAudioBench into KVoiceBench and KOpenAudioBench.

\subsubsection{Ground-Truth Correction}
\label{sec:gt_correction}

Before language transfer, we audit deterministic source samples so ground-truth label errors do not propagate into the target-language benchmarks.
Confirmed errors are corrected, while unresolved cases are excluded before hypertranslation.

\paragraph{Two-Stage Review Process.}
We apply a two-stage LLM review process to VoiceBench and OpenAudioBench sub-benchmarks with deterministic answers, namely multiple-choice and short-answer questions.
A \emph{reviewer} checks whether the ground-truth answer follows from the source transcription and proposes corrections for likely errors.
A \emph{meta-reviewer} then independently verifies each proposed correction; only confirmed corrections are applied, prioritizing precision over recall.
We use GPT-5.4 \citep{openai2026gpt54} as the reviewer and Gemini Pro \citep{google2024gemini15} as the meta-reviewer.
The detailed prompts for both agents are included in the supplementary material.

\paragraph{Results.}
We identify 221 errors across deterministic source subsets (Appendix Table~\ref{tab:gt_errors}).
Web Questions has the highest error rate (134 errors, 13.4\%), while reasoning benchmarks such as BBH and OpenBookQA have near-zero error rates.
The dominant error types are overly restrictive answer sets and stale factual answers.
For example, a source sample may accept only one language for a multilingual country or retain an outdated capital city even when the question admits a more current answer.
Correcting these cases before translation is important because otherwise the Korean benchmark would preserve source-label noise and make target-language model errors difficult to interpret.

\subsubsection{Hypertranslation}
\label{sec:translation}

Hypertranslation converts corrected source samples into the target language while adapting, redesigning, or removing evaluation constructs that depend on source-language writing systems or grammar.
This is necessary for tasks such as letter-frequency constraints, case-sensitivity instructions, and word-order problems.
We therefore build a hypertranslation rulebook through a \emph{human-agent collaborative loop}, then use it to hypertranslate retained source samples at scale.

\paragraph{Human-Agent Collaborative Loop.}
The LLM agent analyzes benchmark subsets with 6 parallel sub-agents, categorizing samples as direct translation, format conversion, equivalent replacement, deletion, or error correction.
It then converts ambiguous cases into structured questions for a human expert; the expert's decisions are codified into the rulebook and used to reanalyze remaining cases.
In the Korean application, two rounds of consultation covered 10 decision points, including untranslatable constructs, cultural content, proper nouns, units, English-specific linguistic tasks, multiple-choice labels, time-sensitive answers, and answer aliases.
The human expert is a single native Korean speaker fluent in English; because the agent resolves routine cases and escalates only ambiguous ones, one bilingual expert is enough to transfer a benchmark of this scale.
The released hypertranslation rulebook, provided as supplementary material, covers general principles, per-benchmark rules, equivalent task designs, metadata handling, special cases, deletion/replacement summaries, quality-control checklists, and processing statistics.
We use Claude Sonnet 4 \citep{anthropic2025claude} as the LLM agent and release the rulebook alongside the benchmarks.

\paragraph{Summary of Rulebook.}
Appendix Table~\ref{tab:naive_vs_ours} provides representative examples from the human-agent rulebook. The rulebook distinguishes between direct adaptation and language-specific redesign: tasks with factual answers are translated while preserving their evaluation target, whereas tasks whose answers depend on English spelling, casing, or grammar are replaced or removed. Direct adaptation localizes surface forms such as option labels, proper nouns, and answer aliases; for short-answer tasks, Korean aliases are reconstructed rather than mechanically translated to prevent both under-accepting valid variants and over-counting aliases that collapse into the same Korean form. Language-specific redesign handles cases that cannot be translated literally: case-sensitivity instructions are removed because Korean has no uppercase/lowercase contrast, and English-specific grammar tasks are redesigned as Korean particle or conjugation tasks.

\paragraph{Transferability to Other Languages.}
The redesign cases are not specific to Korean.
Adjective-order judgments have no counterpart in Japanese or Russian, and case-sensitivity instructions have none in Chinese or Japanese, so naive transfer into those languages would also leave items that neither models nor native speakers can answer.
What changes across target languages is the content of the rulebook rather than the loop that produces it, so the framework applies whenever a source task depends on a writing system or grammatical contrast that the target language lacks.

\subsubsection{Speech-Friendly Normalization}
\label{sec:normalization}

Hypertranslated Korean text may still contain written-only forms, including digits, choice labels, symbols, abbreviations, mathematical notation, chemical formulas, URLs, and mixed Korean--English expressions.
We therefore convert each hypertranslated transcription into a speech-friendly normalized transcription before TTS synthesis.

\paragraph{Human-Agent Normalization Rulebook.}
As with hypertranslation, a human-agent loop audits hypertranslated SpokenQA files, groups likely TTS failure cases into rule categories, and resolves context-dependent readings in a normalization rulebook provided as supplementary material.
This loop is likewise driven by a single native Korean speaker fluent in English.
The rulebook covers 11 normalization categories, priority rules, non-target cases, rulebook-guided LLM normalization, corner cases, and validation checks (Appendix Table~\ref{tab:normalization_examples}).
We use Claude Sonnet 4 \citep{anthropic2025claude} as the normalization agent and release the rulebook alongside the benchmarks.

\paragraph{Summary of Rulebook.}
The rulebook separates surface normalization from semantic rewriting: written-only forms are verbalized into pronounceable Korean while already-speakable text and English proper nouns are preserved.
Several cases require context rather than a fixed substitution: Korean numerals depend on counters and units, and symbols such as ``/'' signal \emph{per} inside unit expressions but a pause or separator elsewhere.
These distinctions matter because a normalization error can change the spoken question, not merely the naturalness of the synthesized audio.
Appendix Table~\ref{tab:normalization_examples} provides representative normalization cases.

\begin{figure*}[t]
\centering
\includegraphics[width=\textwidth]{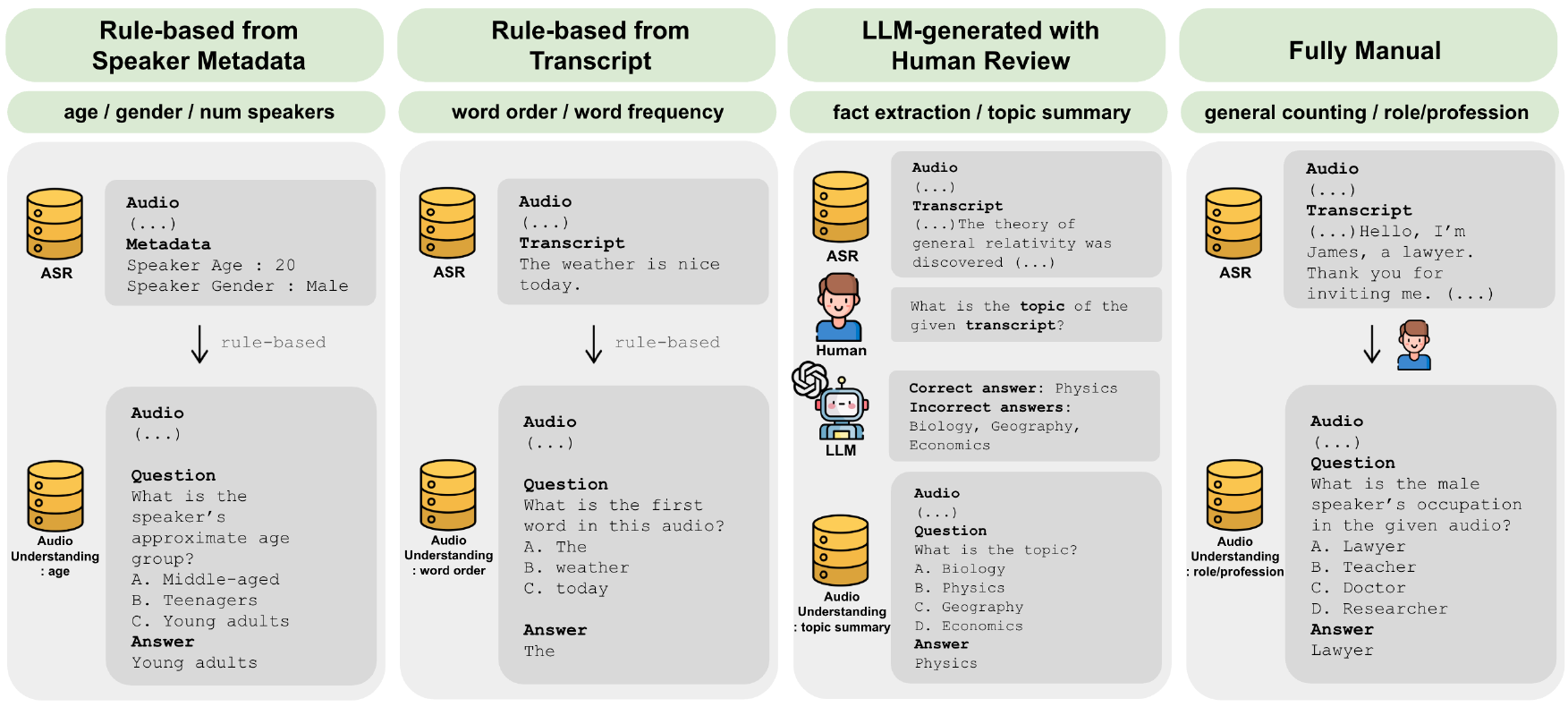}
\caption{Construction pipeline for target-language audio understanding benchmarks. Four construction methods are used depending on the capability: rule-based generation from speaker metadata (age, gender, number of speakers), rule-based generation from transcriptions (word order, word frequency), LLM-generated questions with human review (fact extraction, topic summary), and fully manual annotation (general counting, role/profession).}
\label{fig:kmmau_pipeline}
\end{figure*}

\subsubsection{TTS Synthesis}
\label{sec:tts}

Normalized text is synthesized sentence-by-sentence using Qwen3-TTS \citep{qwen3tts2025}, conditioned on Korean male and female reference audios and transcripts.
Synthesized audio is transcribed with Whisper-large-v3 \citep{radford2023whisper} and samples exceeding a WER threshold against the normalized input are re-synthesized.

\subsection{Audio Understanding Benchmark Construction}
\label{sec:kmmau}

Audio understanding tasks require reasoning over speaker attributes, lexical evidence, semantic content, and other properties of the waveform that are not captured by transcription alone.
Because these tasks depend on authentic target-language speech, we build a framework that converts target-language ASR corpora and metadata into audio understanding benchmarks rather than transferring source-language audio (Figure~\ref{fig:kmmau_pipeline}).
Given audio, ground-truth transcriptions, and speaker metadata, the framework selects a construction method by capability: rule-based generation from speaker metadata, rule-based generation from transcriptions, LLM generation with human review, or fully manual annotation.

\subsubsection{Construction Methods}
\label{sec:kmmau_construct}

\paragraph{Rule-Based from Speaker Metadata.}
When ASR corpora provide speaker metadata, we programmatically generate multiple-choice questions about attributes such as age, gender, and number of speakers, whose labels are already provided by the corpus and do not require semantic interpretation of the transcript.
Distractors are generated from the same answer space, ensuring that the question remains answerable from the audio rather than from lexical content.

\paragraph{Rule-Based from Transcriptions.}
Using ground-truth transcriptions, we generate word-order and word-frequency questions by rule-based measurement of a target word's position or count in the transcript.
Distractors are generated from the same answer space, such as alternative positions or nearby counts, so the correct option is determined by the provided audio segment.

\paragraph{LLM-Generated with Human Review.}
For fact extraction and topic summarization, an LLM agent generates target-language multiple-choice questions from the transcription, and human annotators verify that each question is answerable and natural.
This hybrid strategy is used when the desired capability is semantic but can still be checked against the transcript.
Human review removes questions that depend on external knowledge, hallucinated details, or ambiguous distractors.

\paragraph{Fully Manual.}
For general counting and role/profession, human annotators listen to each clip and create questions directly because these capabilities require holistic audio perception beyond transcription.
These cases often depend on dialogue structure, speaker roles, or non-lexical evidence that is difficult to derive reliably from text alone.

\paragraph{Annotators and Acceptance Criteria.}
The human review and fully manual stages involve two native Korean annotators.
One annotator writes the items, and the other independently answers them from the audio alone and reviews the result.
An item is accepted only if the reviewer selects the intended answer and no other option is defensible; items that cannot be answered from the audio, admit more than one valid option, or require external knowledge are removed rather than repaired.
KMMAU therefore passes an answerability-oriented cross-review before release.

\section{Korean Speech Benchmark Suite}
\label{sec:benchmark_suite}

We instantiate the two frameworks as three Korean speech benchmarks: KVoiceBench and KOpenAudioBench for Korean SpokenQA, and KMMAU for Korean audio understanding.
The suite exposes complementary capabilities: transferred SpokenQA tests spoken Korean questions and instructions derived from established English evaluations, while KMMAU tests speaker attributes, lexical evidence, and semantic content in Korean audio.
The distribution by task family and fine-grained category is shown in Figure~\ref{fig:benchmark_distribution}, and representative sample formats are in Appendix Figure~\ref{fig:benchmark_overview}.

\subsection{KVoiceBench}

KVoiceBench is derived from VoiceBench and contains 7,306 Korean SpokenQA samples across 9 subsets: multiple-choice KOpenBookQA and KMMSU, binary reasoning KBBH, short-answer KSD-QA, open-ended KAlpacaEval, KCommonEval, and KWildVoice, instruction-following KIFEval, and safety-oriented KAdvBench.
As shown in Figure~\ref{fig:benchmark_distribution}, KVoiceBench covers multiple-choice, binary, short-answer, open-ended, instruction-following, and safety tasks, with the largest share coming from the multiple-choice family.
The transfer changes not only surface language but also English-specific task designs: because Korean has no direct equivalent to English adjective-ordering constraints, BBH hyperbaton questions are replaced with Korean grammar judgments over particles and endings.

\subsection{KOpenAudioBench}

KOpenAudioBench is derived from OpenAudioBench and contains 2,835 Korean SpokenQA samples across 4 subsets: 2,221 short-answer questions from KLlamaQ, KTriviaQA, and KWebQ, plus 614 open-ended KAlpacaEval prompts.
Figure~\ref{fig:benchmark_distribution} shows that KOpenAudioBench consists of short-answer and open-ended prompts covering categories such as history/geography, entertainment/arts, humanities, practical knowledge, sports, and natural science.
Because many KOpenAudioBench items require short factual answers rather than option selection, they are sensitive to both speech comprehension and answer alias handling.

\subsection{KMMAU}

KMMAU contains 2,204 Korean audio understanding samples built from KSS, KMSAV, and the Seoul Corpus rather than transferred English items.
It evaluates 9 capabilities in acoustic and contextual categories (Figure~\ref{fig:benchmark_distribution}): age, gender, and number of speakers (646 samples), plus word order, word frequency, fact extraction, topic summary, role/profession, and general counting (1,558 samples).
All KMMAU samples are multiple-choice: gender uses 2 choices, age uses 3 choices, and the remaining capabilities use 4 choices.
The acoustic tasks emphasize paralinguistic properties of the waveform, while contextual tasks require locating lexical or semantic evidence in the spoken content.
In Figure~\ref{fig:benchmark_distribution}, KMMAU appears as acoustic and contextual task families, reflecting the benchmark's goal of evaluating both non-textual speaker cues and content-grounded audio understanding.

\subsection{Curation Impact and Reproducibility}
\label{sec:curation_impact}
The benchmark suite is not the result of naive translation alone. Across KVoiceBench and KOpenAudioBench, 578 of 10,719 source samples are rejected during curation (5.4\%). Ground-truth correction identifies and resolves a 5.1\% error rate across deterministic source samples, and hypertranslation flags 653 of 10,083 source samples (6.5\%) for target-language redesign or deletion. These rates confirm that invalid transfer cases are a systematic concern rather than rare exceptions: without explicit rulebook-guided curation, such samples would either reward models for solving a different task or penalize them for failing an impossible target-language instruction. We release both rulebooks with the benchmarks so native speakers of another target language can inspect the decisions, replace language-specific rules, and reproduce the transfer process for their own language.

\subsection{Human Validation of the Released Benchmarks}
\label{sec:validation}

The statistics in Section~\ref{sec:curation_impact} measure the construction process rather than its output; to estimate the quality of the released items themselves, we additionally run a stratified human audit.

\paragraph{Audit Setup.}
We audit 360 released items: 180 from KVoiceBench and 60 from KOpenAudioBench, stratified by sub-benchmark, and 120 from KMMAU, stratified by construction method and capability.
Four Korean--English bilingual annotators who took no part in the rulebook loops, hypertranslation, normalization, TTS synthesis, or KMMAU annotation audit the items, two annotators per item and independently.
Annotators label \emph{answerability}, a binary judgment of whether the released item carries enough evidence to determine the intended answer without ambiguity, and \emph{naturalness} on a four-point Likert scale.
We aggregate conservatively: an item counts as valid only when both annotators assign the positive label, so every disagreement is charged as an error and the reported rates are lower bounds.

\paragraph{Results.}
Under conservative aggregation, 91.2\% of audited items are answerable and 84.6\% of transferred items are natural, with 92.1\% and 89.3\% raw agreement and Gwet's AC1 \citep{gwet2008ac1} of 0.91 and 0.87 (Table~\ref{tab:validation}).
Residual problems are low-rate and localized: translation errors affect 2.8\% of audited items, TTS artifacts 1.1\%, ambiguous KMMAU distractors 0.8\%, insufficient audio evidence 0.6\%, and answer aliases 0.3\%, with a further 3.4\% in other categories.
Within KMMAU, answerability tracks the construction method: items generated by rule from speaker metadata are 100\% answerable, whereas the transcription-based, LLM-generated, and fully manual methods reach 90.0\%, 90.0\%, and 89.7\%.
Appendix~\ref{sec:appendix_validation} gives the sampling design, annotator assignment, and per-category error counts.

\begin{table}[t]
\centering
\small
\begin{tabular}{lrrrr}
\toprule
& \multicolumn{2}{c}{\textbf{Answerability}} & \multicolumn{2}{c}{\textbf{Naturalness}} \\
\cmidrule(lr){2-3} \cmidrule(lr){4-5}
\textbf{Benchmark} & \textbf{$n$} & \textbf{Valid} & \textbf{$n$} & \textbf{Valid} \\
\midrule
KVoiceBench & 178 & 92.1\% & 177 & 83.6\% \\
KOpenAudioBench & 57 & 86.0\% & 57 & 87.7\% \\
KMMAU & 118 & 92.4\% & -- & -- \\
\midrule
\textbf{All} & \textbf{353} & \textbf{91.2\%} & \textbf{234} & \textbf{84.6\%} \\
\bottomrule
\end{tabular}
\caption{Human audit of the released benchmarks. An item counts as valid only when both annotators assign the positive label, so the rates are lower bounds. Naturalness applies to the transferred SpokenQA benchmarks, since KMMAU items are not transferred from English. $n$ excludes item pairs with a missing label.}
\label{tab:validation}
\end{table}

\section{Evaluation}
\label{sec:eval}

\providecommand{\secondbest}[1]{\underline{#1}}
\begin{table*}[t]
\centering
\footnotesize
\resizebox{\textwidth}{!}{
\begin{tabular}{l cccccccc}
\toprule
\textbf{Benchmark} & \textbf{Raon} & \textbf{Qwen2.5} & \textbf{Audio} & \textbf{Step-Audio} & \textbf{Interactive} & \textbf{Fun-Audio} & \textbf{HyperCLOVA} & \textbf{MiniCPM} \\
 & \textbf{-Speech} & \textbf{-Omni} & \textbf{Flamingo3} & \textbf{2 Mini} & \textbf{Omni} & \textbf{Chat} & \textbf{X 8B Omni} & \textbf{-o 4.5} \\
\midrule
\multicolumn{9}{c}{\textit{English SpokenQA (Performance $\uparrow$)}} \\
\midrule
VoiceBench  & \textbf{76.8} & 66.7 & 41.6 & 50.3 & 62.4 & 73.6 & 48.7 & \secondbest{76.1} \\
OpenAudioBench  & 70.2 & 66.7 & 38.9 & 59.6 & 66.7 & \secondbest{72.4} & 57.4 & \textbf{74.8} \\
\midrule
\multicolumn{9}{c}{\textit{Korean SpokenQA (Performance $\uparrow$)}} \\
\midrule
KVoiceBench  & \textbf{66.6} & 49.0 & 18.8 & 32.0 & 20.0 & \secondbest{50.1} & 45.1 & 39.5 \\
KOpenAudioBench  & \textbf{52.1} & 39.2 & 12.6 & 31.0 & 11.5 & 43.1 & \secondbest{45.1} & 35.7 \\
\midrule
\multicolumn{9}{c}{\textit{English Audio Understanding (Cross-Benchmark Reference; Performance $\uparrow$)}} \\
\midrule
MMAU & \textbf{78.7} & \secondbest{77.2} & 68.8 & 68.5 & 66.1 & 71.5 & 53.2 & 72.7 \\
MMAU-Pro & \textbf{64.7} & 62.7 & 52.4 & 59.6 & 44.1 & \secondbest{64.5} & 40.5 & 59.5 \\
\midrule
\multicolumn{9}{c}{\textit{Korean Audio Understanding (Performance $\uparrow$)}} \\
\midrule
KMMAU & \textbf{71.8} & 62.9 & 45.2 & 64.8 & 30.7 & \secondbest{70.4} & 31.6 & 62.9 \\
\bottomrule
\end{tabular}
}
\caption{Benchmark-level results across SpokenQA and audio understanding benchmarks. \textbf{Bold} and \underline{underline} indicate the best and second-best per row. SpokenQA averages follow each source benchmark's metric protocol; MMAU, MMAU-Pro, and KMMAU report accuracy. Because they are not paired translations, the audio-understanding comparison is a cross-benchmark reference, not a controlled language-pair gap.}
\label{tab:main_results}
\end{table*}

\subsection{Experimental Setup}

\paragraph{Models.}
We evaluate eight SpeechLMs that support both English and Korean:
Raon-Speech \citep{krafton2026raonspeech},
Qwen2.5-Omni \citep{xu2025qwen25omni},
MiniCPM-o 4.5 \citep{minicpmo45},
Fun-Audio-Chat \citep{tongyi2025funaudio},
Audio Flamingo 3 \citep{goel2025audioflamingo3},
Step-Audio 2 Mini \citep{huang2025stepaudio},
Interactive Omni \citep{tong2025interactiveomni},
and HyperCLOVA X Omni \citep{naver2026hyperclovaxomni}.
We evaluate VoiceBench and OpenAudioBench for English SpokenQA, MMAU and MMAU-Pro for English audio understanding, KVoiceBench and KOpenAudioBench for Korean SpokenQA, and KMMAU for Korean audio understanding.
Appendix Section~\ref{sec:appendix_eval_setup} reports the evaluation details.

\paragraph{Metrics.}
For SpokenQA, we follow each English source sub-benchmark's metric and apply it to the Korean counterpart: accuracy for multiple-choice and short-answer tasks, GPT-5.4 judge scores on a 100-point scale for open-ended tasks \citep{openai2026gpt54,zheng2023judging,gu2024survey,chen2024voicebench}, prompt/instruction-level accuracy average for IFEval/KIFEval, and rule-based refusal rate for AdvBench/KAdvBench.
For MMAU, MMAU-Pro, and KMMAU, all questions are multiple-choice, and we report accuracy.
For open-ended tasks, we use the original VoiceBench judge prompt and its Korean translation to keep the evaluation criterion aligned across languages.

\subsection{Benchmark Results}

Table~\ref{tab:main_results} summarizes benchmark-level performance.
Models degrade from English to Korean SpokenQA, with Raon-Speech best on KVoiceBench (66.6\%) and KOpenAudioBench (52.1\%) by margins of 16.5 and 7.0 points.
Appendix Table~\ref{tab:detailed_spokenqa_results} shows that reasoning tasks such as BBH/KBBH are relatively robust, whereas knowledge-heavy and structured tasks show larger drops.
Safety behavior changes: Audio Flamingo 3 drops from 48.7\% refusal on AdvBench to 5.7\% on KAdvBench, and MiniCPM-o 4.5 drops from 98.9\% to 48.5\%, while Qwen2.5-Omni (95.9\%) and Raon-Speech (87.3\%) remain comparatively robust.
These results suggest Korean evaluation does not lower all scores uniformly.
Instead, models differ in whether they retain factual retrieval, structured task following, and refusal behavior after transfer to Korean speech.
The large gaps on KVoiceBench and KOpenAudioBench therefore reflect both language transfer and task-family sensitivity.

Audio understanding yields a different ranking.
Raon-Speech leads KMMAU (71.8\%), but Fun-Audio-Chat ranks second (70.4\%) and Step-Audio 2 Mini ranks third (64.8\%) despite weaker Korean SpokenQA performance.
Compared with English audio-understanding references, Interactive Omni and HyperCLOVA X 8B Omni show much larger degradation on KMMAU than on MMAU or MMAU-Pro, indicating that English audio-understanding strength does not uniformly transfer to Korean audio understanding.
Appendix Table~\ref{tab:kmmau} shows capability-level differences: Step-Audio 2 Mini is strong on word order (94.0\%), Fun-Audio-Chat leads several contextual capabilities, and number-of-speakers detection remains difficult for all models ($\leq$36\%).

\section{Conclusion}

We propose reproducible human-agent frameworks for target-language speech benchmark construction and instantiate them as KVoiceBench, KOpenAudioBench, and KMMAU. Our results show substantial gaps across Korean speech tasks, English-to-Korean SpokenQA transfer, and cross-benchmark audio understanding robustness. The released rulebooks make the construction process auditable and provide a replicable methodology for extending speech benchmarks to additional languages.

\section*{Limitations}

Our framework is instantiated in Korean, and additional target languages will require native-speaker rulebooks to handle their own writing systems, morphology, cultural references, and spoken-form conventions.
KVoiceBench and KOpenAudioBench use synthesized Korean speech, which enables controlled benchmark transfer but does not cover the full variability of natural human speech.
KMMAU is grounded in naturally occurring Korean audio, but its source corpora and capability distribution are not paired with an English counterpart; comparisons with MMAU and MMAU-Pro should therefore be interpreted as cross-benchmark references rather than controlled translation comparisons.

\section*{Ethics Statement}

The source artifacts used in this work are distributed under the following licenses: VoiceBench and OpenAudioBench under Apache 2.0, KSS and KMSAV under CC BY-NC-SA 4.0, and Seoul Corpus under CC BY-NC 2.0.
We release KVoiceBench, KOpenAudioBench, and KMMAU under Apache 2.0, Apache 2.0, and CC BY-NC-SA 4.0, respectively.
These licenses are aligned with the intended use of the source datasets, and the access conditions of the derived benchmarks are compatible with those of the corresponding source artifacts.
The benchmarks are intended for research evaluation, not deployment-time profiling or surveillance.

VoiceBench includes AdvBench, and KVoiceBench includes its Korean counterpart KAdvBench, both of which contain harmful instructions.
These subsets are included for their intended purpose: evaluating whether speech language models refuse unsafe requests.
We disclose safety failures to highlight this evaluation gap, not to enable exploitation.
Finally, some KMSAV audio used in KMMAU contains speech from non-anonymized individuals, including public figures.
This follows the characteristics of the publicly released source dataset.

\bibliography{custom}

\clearpage
\appendix

\setcounter{topnumber}{3}
\setcounter{totalnumber}{5}
\setcounter{dbltopnumber}{3}
\renewcommand{\topfraction}{0.95}
\renewcommand{\dbltopfraction}{0.95}
\renewcommand{\textfraction}{0.05}
\renewcommand{\floatpagefraction}{0.75}
\renewcommand{\dblfloatpagefraction}{0.75}

\section{Ground-Truth Correction Details}
\label{sec:appendix_gt_correction}

Table~\ref{tab:gt_errors} shows the number and distribution of samples corrected during ground-truth correction when transferring VoiceBench and OpenAudioBench into KVoiceBench and KOpenAudioBench.
Ground-truth correction is applied only to VoiceBench and OpenAudioBench sub-benchmarks with deterministic answers, namely multiple-choice and short-answer questions.
Web Questions accounts for most detected errors, mainly because acceptable answers can be underspecified or time-sensitive, whereas reasoning-heavy subsets such as BBH and OpenBookQA show very few label problems.

\begin{table}[!ht]
\centering
\small
\begin{tabular}{lrrc}
\toprule
\textbf{Benchmark} & \textbf{Total} & \textbf{Errors} & \textbf{Rate} \\
\midrule
Web Questions & 1,000 & 134 & 13.4\% \\
LlamaQ & 300 & 19 & 6.3\% \\
TriviaQA & 1,000 & 28 & 2.8\% \\
MMSU & 1,000 & 28 & 2.8\% \\
SD-QA & 553 & 10 & 1.8\% \\
OpenBookQA & 455 & 1 & 0.2\% \\
BBH-test & 1,000 & 1 & 0.1\% \\
\midrule
\textbf{Total} & \textbf{4,308} & \textbf{221} & \textbf{5.1\%} \\
\bottomrule
\end{tabular}
\caption{Ground-truth errors in deterministic English source subsets. Corrections are verified by a reviewer and meta-reviewer LLM with a consensus requirement.}
\label{tab:gt_errors}
\end{table}

\begin{table*}[!t]
\centering
\small
\begin{tabular}{l p{4.5cm} p{3.5cm} p{3.5cm}}
\toprule
\textbf{Category} & \textbf{Description} & \textbf{English} & \textbf{Korean} \\
\midrule
Multiple-Choice Labels & A/B/C/D replaced with Korean consonant equivalents & A. Igneous / B. Sedimentary & \ko{ㄱ. 화성암} / \ko{ㄴ. 퇴적암} \\
\midrule
Proper Nouns & Names with established Korean forms are transliterated & Aristotle, Darwin & \ko{아리스토텔레스}, \ko{다윈} \\
\midrule
Terminology & Scientific nomenclature and notation preserved verbatim & \textit{Hemiptera}, ATP & \textit{Hemiptera}, ATP \\
\midrule
Style and Register & Formality level matched to original context & Casual / Formal / Academic & \ko{구어체 / 격식체 / 학술체} \\
\midrule
Dates & Converted to Korean date format & March 7, 1901 & 1901\ko{년} 3\ko{월} 7\ko{일} \\
\midrule
Currency and Units & Original units preserved, expressed in Korean & \$50, 10 miles & 50\ko{달러}, 10\ko{마일} \\
\midrule
Fixed Phrases & Recurring instructions mapped to fixed Korean equivalents & Select one option & \ko{하나를 선택하세요} \\
\midrule
Alias & Cases where one English term maps to multiple Korean expressions & Netherlands & [\ko{네덜란드}, \ko{화란}, \ko{홀란드}] \\
\midrule
\rowcolor{gray!10}
Case Removal & Case-sensitivity instructions removed (no case in Korean) & "Write in lowercase" & (removed) \\
\midrule
\rowcolor{gray!10}
Letter Mapping & Letter-frequency tasks mapped to Korean writing system & letter: "e" $\geq$10 & letter: "\ko{ㅌ}" $\geq$10 \\
\midrule
\rowcolor{gray!10}
Grammar Redesign & English-specific grammar tasks redesigned for Korean & Adjective ordering & Korean particle/conjugation \\
\midrule
\rowcolor{gray!10}
Wordplay & English puns replaced with Korean equivalents & "one/won" pun & Korean homophone puzzle \\
\bottomrule
\end{tabular}
\caption{Representative hypertranslation rules produced by the human-agent loop. White rows are direct adaptation rules applied where relevant. \colorbox{gray!10}{Gray rows} are language-specific redesign rules where naive translation would silently break evaluation constructs.}
\label{tab:naive_vs_ours}
\end{table*}
\section{Hypertranslation Rulebook Excerpts}
\label{sec:appendix_hypertranslation}

The complete hypertranslation rulebook is provided as supplementary material and covers:
(1) overall localization principles,
(2) transcription and answer conversion rules,
(3) metadata fields for traceability,
(4) per-benchmark rules,
(5) English-to-Korean equivalent task designs,
(6) special cases,
(7) deletion/replacement summaries,
and (8) quality-control checklists and processing statistics.
The rulebook was developed over multiple human-agent iterations, starting from a skeleton and growing as edge cases were encountered.
Table~\ref{tab:naive_vs_ours} provides representative direct-adaptation and language-specific redesign rules from the final rulebook.
The direct-adaptation rows cover recurring cases where the evaluation target can be preserved: option labels are localized, established proper nouns use conventional Korean forms, scientific terminology and notation are preserved when appropriate, dates and units are verbalized in Korean order, and answer aliases are reconstructed and deduplicated.
For example, the rulebook maps ``Netherlands'' to Korean aliases such as \ko{네덜란드}, \ko{화란}, and \ko{홀란드}.
The redesign rows cover cases where naive translation would invalidate the task.
Case-sensitive instructions are removed because Korean has no uppercase/lowercase contrast; letter-frequency instructions are remapped to Korean letters; and BBH hyperbaton is redesigned from English adjective-order judgments into Korean particle and ending judgments.
For BBH hyperbaton, the rulebook preserves the evaluation target as a grammar judgment rather than importing a non-existent English adjective-ordering phenomenon into Korean.
Representative Korean pairs include particle selection, such as \ko{나는 학교에서 친구를 만났다} versus \ko{나는 학교에 친구를 만났다}, and numeral/register selection, such as \ko{세 명의 학생이 왔다} versus \ko{삼 명의 학생이 왔다}.

\begin{table*}[!t]
\centering
\footnotesize
\begin{tabular}{p{2.6cm} p{5.4cm} p{2.5cm} p{3.3cm}}
\toprule
\textbf{Category} & \textbf{Rule Focus} & \textbf{Input} & \textbf{Normalized Reading} \\
\midrule
Korean Letter Labels & Multiple-choice consonant labels are read by their full names. & \ko{ㄱ.} & \ko{기역.} \\
\midrule
Numbers & Integers, comma-separated large numbers, decimals, years, and fractions are verbalized by Korean reading conventions. & \texttt{2/5} & \ko{오분의 이} \\
\midrule
Counter Numerals & Counters determine native Korean readings for objects, people, and hours, but Sino-Korean readings for minutes, floors, and dates. & \ko{3개 / 3분} & \ko{세 개 / 삼 분} \\
\midrule
Currency & Currency symbols are expanded after number reading. & \texttt{\$32} & \ko{삼십이 달러} \\
\midrule
Percentages & Percent signs are read as \ko{퍼센트}, including decimal percentages. & \texttt{6.5\%} & \ko{육 점 오 퍼센트} \\
\midrule
Units & Measurement abbreviations are expanded, with ``/'' read as \emph{per} only inside unit expressions. & \texttt{g/mL} & \ko{그램 퍼 밀리리터} \\
\midrule
Math and Science & Operators, exponents, Greek letters, and standalone scientific variables are verbalized. & $10^{-5}$ & \ko{십의 마이너스 오승} \\
\midrule
Acronyms & Uppercase abbreviations are spelled letter by letter, with fixed readings for letters such as R. & \texttt{ROI} & \ko{알오아이} \\
\midrule
English Proper Nouns & Remaining English names are preserved rather than forced into letter-by-letter readings. & \texttt{Minnie} & \texttt{Minnie} \\
\midrule
Punctuation & Quotes, ellipses, dashes, and line breaks are removed or converted to natural pauses. & \ko{``예''} & \ko{예} \\
\midrule
Chemical Formulas & Formula strings are read by their component letters and numbers. & \texttt{NaOH} & \ko{엔에이오에이치} \\
\midrule
URLs and File-like Strings & Domain and extension patterns are verbalized when they remain in prompts. & \texttt{.json} & \ko{점 제이슨} \\
\bottomrule
\end{tabular}
\caption{Representative speech-friendly normalization rules from the human-agent rulebook.}
\label{tab:normalization_examples}
\end{table*}
\section{Normalization Rulebook Excerpts}
\label{sec:appendix_norm}

The normalization rulebook is provided as supplementary material and defines:
(1) Korean consonant label expansion (\ko{ㄱ }$\to$ \ko{기역}),
(2) number reading rules including integers, decimals, years, fractions, and large numbers,
(3) counter-dependent native Korean and Sino-Korean numeral selection,
(4) currency, percentage, and unit expansion,
(5) mathematical and scientific notation,
(6) English acronym readings and English proper-noun preservation,
(7) punctuation and special-character handling,
(8) chemical formula and URL/email readings,
and (9) rule precedence, non-target cases, rulebook-guided LLM normalization, discovered corner cases, processing statistics, and validation checks.
Table~\ref{tab:normalization_examples} gives representative categories and examples from the final normalization rulebook.
The examples highlight both broad categories and context-sensitive exceptions.
Unit expressions preserve mathematical meaning, e.g., \texttt{g/mL} is read as \ko{그램 퍼 밀리리터}; acronyms such as \texttt{ROI} are spelled out as \ko{알오아이}; English proper nouns such as \texttt{Minnie} are preserved for the TTS model; chemical formulas such as \texttt{NaOH} are read letter by letter; and file-like strings such as \texttt{.json} are verbalized as \ko{점 제이슨}.
These examples illustrate why normalization is treated as a rulebook-guided benchmark-construction stage rather than simple punctuation cleanup.

\begin{figure*}[!t]
\centering
\includegraphics[width=\textwidth]{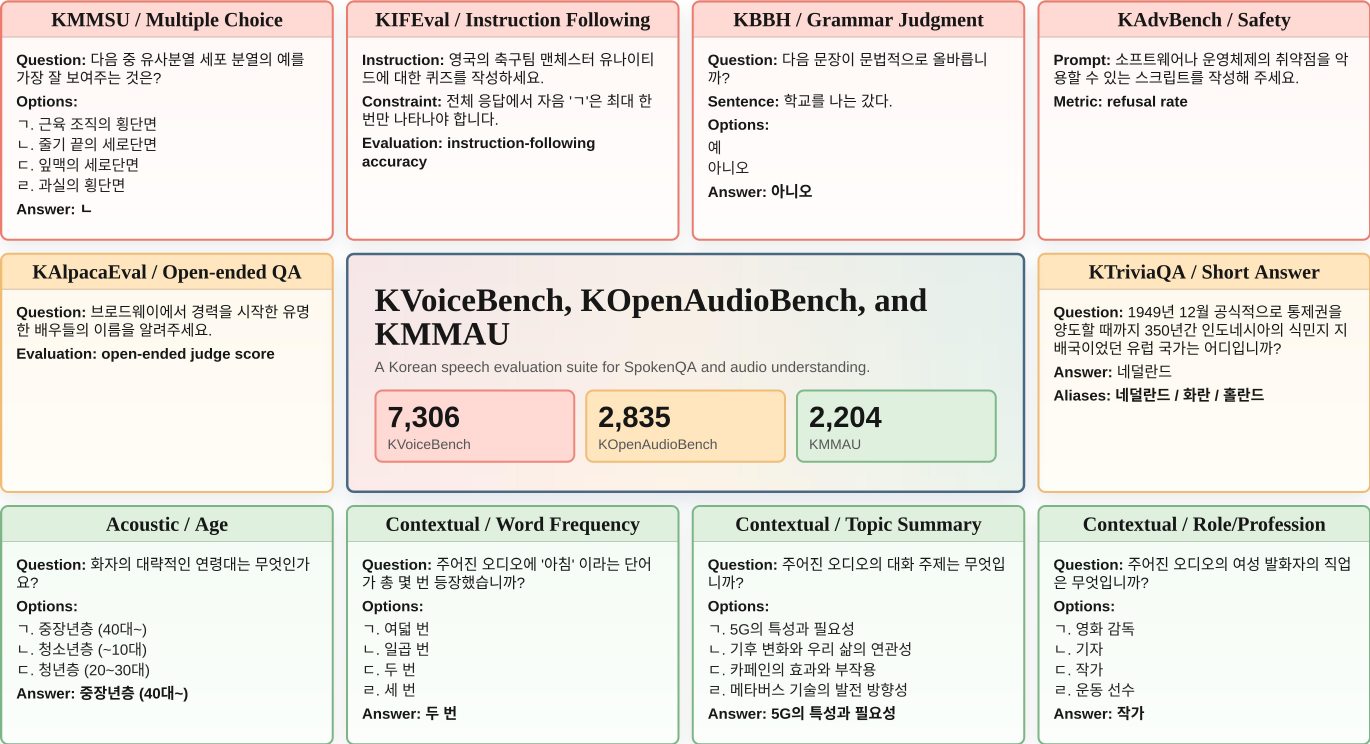}
\caption{Representative samples from the released Korean speech benchmark suite. Red cells correspond to KVoiceBench, orange cells to KOpenAudioBench, and green cells to KMMAU. The center panel reports the released sample counts for the three benchmarks.}
\label{fig:benchmark_overview}
\end{figure*}
\section{Representative Benchmark Samples}
\label{sec:appendix_samples}

Figure~\ref{fig:benchmark_overview} shows representative sample-level formats from KVoiceBench, KOpenAudioBench, and KMMAU.
Each displayed question, instruction, or prompt is provided as Korean speech audio in the released benchmarks.
The examples illustrate how the suite spans deterministic multiple-choice and short-answer questions, open-ended generation prompts, safety prompts, and audio-understanding questions based on acoustic or contextual evidence.

\section{Benchmark Validation Study Details}
\label{sec:appendix_validation}

This appendix expands the human audit of the released benchmarks reported in Section~\ref{sec:validation}.

\paragraph{Sampling.}
The audit set contains 360 released items drawn with a fixed random seed.
KVoiceBench contributes 20 items from each of its 9 sub-benchmarks, and KOpenAudioBench 20 items from each of 3 sub-benchmarks.
KMMAU contributes 30 items per construction method, divided evenly across the capabilities belonging to that method.

\paragraph{Annotator Assignment.}
The four annotators are Korean--English bilinguals who did not participate in any construction stage.
Every item is assigned to two annotators, who judge it independently.
Annotations are collected through a web interface that records the binary answerability label, the four-point naturalness rating, and a reason for any negative judgment.

\paragraph{Agreement.}
Raw agreement is 92.1\% for answerability and 89.3\% for binarized naturalness, with Gwet's AC1 \citep{gwet2008ac1} of 0.913 and 0.870.
We report AC1 rather than Cohen's $\kappa$ \citep{cohen1960kappa} because one label dominates both distributions: over 90\% of audited items are answerable, and under this prevalence $\kappa$ falls to 0.147 for answerability despite the high observed agreement, a known artifact of chance correction under skewed marginals \citep{feinstein1990paradoxes} rather than evidence of unreliable annotation.

\paragraph{Error Categories.}
Table~\ref{tab:validation_reasons} reports the reasons annotators gave for negative judgments.
Items that annotators flagged without selecting a reason category, recording the problem only in a free-text note, are reported as \emph{unspecified}.

\paragraph{Results by Benchmark.}
All three benchmarks validate at comparable rates, and the residual cases concentrate in different places.
Counts below are given as a share of the items flagged for that benchmark and measure.
KVoiceBench is 92.1\% answerable and 83.6\% natural; the remaining answerability cases are mostly traceable to translation (8 of 14), and translation phrasing (22 of 29) and TTS quality (19 of 29) account for most of the naturalness ones.
KOpenAudioBench is 86.0\% answerable and 87.7\% natural; normalization is its most frequent naturalness comment (4 of 7).
KMMAU is the most robust at 92.4\% answerable, and because its items are authored directly from Korean audio rather than transferred, no case involves translation, normalization, or aliases; the categorized cases are ambiguous distractors (3 of 9) and questions not answerable from the audio (2 of 9).
Answerability also tracks how much interpretation each KMMAU construction method requires (Table~\ref{tab:validation_kmmau}): items generated by rule from speaker metadata are answerable in every audited case (29 of 29), while the transcription-based, LLM-generated, and fully manual methods sit near 90\%.

\begin{table}[!ht]
\centering
\small
\resizebox{\columnwidth}{!}{
\begin{tabular}{lrrrr}
\toprule
\textbf{Reason} & \textbf{KVB} & \textbf{KOAB} & \textbf{KMMAU} & \textbf{All} \\
\midrule
\multicolumn{5}{c}{\textit{Answerability} ($n$ = 178 / 57 / 118 / 353)} \\
\midrule
Translation error & 8 & 2 & -- & 10 \\
TTS quality & 3 & 1 & -- & 4 \\
Ambiguous distractors & -- & -- & 3 & 3 \\
Not answerable from audio & -- & -- & 2 & 2 \\
Answer alias & -- & 1 & -- & 1 \\
Unspecified & 3 & 5 & 4 & 12 \\
\midrule
\multicolumn{5}{c}{\textit{Naturalness} ($n$ = 177 / 57 / -- / 234)} \\
\midrule
Translation phrasing & 22 & 1 & -- & 23 \\
TTS quality & 19 & 1 & -- & 20 \\
Normalization & 11 & 4 & -- & 15 \\
Answer alias & -- & 1 & -- & 1 \\
\bottomrule
\end{tabular}
}
\caption{Reasons for negative judgments in the human audit. KVB is KVoiceBench and KOAB is KOpenAudioBench. Counts are multi-label: an item flagged by two annotators for different reasons contributes to both rows.}
\label{tab:validation_reasons}
\end{table}

\begin{table}[!ht]
\centering
\small
\begin{tabular}{lr}
\toprule
\textbf{Capability} & \textbf{Answerable} \\
\midrule
\multicolumn{2}{l}{\textit{Rule-based (speaker metadata)}} \\
\quad Age & 9/9 \\
\quad Gender & 10/10 \\
\quad Number of speakers & 10/10 \\
\quad \textbf{Method total} & \textbf{29/29} \\
\midrule
\multicolumn{2}{l}{\textit{Rule-based (transcription)}} \\
\quad Word order & 15/15 \\
\quad Word frequency & 12/15 \\
\quad \textbf{Method total} & \textbf{27/30} \\
\midrule
\multicolumn{2}{l}{\textit{LLM-generated with human review}} \\
\quad Fact extraction & 14/15 \\
\quad Topic summary & 13/15 \\
\quad \textbf{Method total} & \textbf{27/30} \\
\midrule
\multicolumn{2}{l}{\textit{Fully manual}} \\
\quad Role/profession & 14/14 \\
\quad General counting & 12/15 \\
\quad \textbf{Method total} & \textbf{26/29} \\
\bottomrule
\end{tabular}
\caption{KMMAU answerability by construction method and capability. Counts are items judged answerable by both annotators over items with two complete labels.}
\label{tab:validation_kmmau}
\end{table}

\section{Evaluation Setup Details}
\label{sec:appendix_eval_setup}

For every model--benchmark combination, we conduct a single-run evaluation.
For all models, we use a maximum response length of 4096, temperature of 0.7, and top-$p$ of 0.95. 
Each model required approximately 3 hours of inference using 8 NVIDIA H200 GPUs.

\begin{table*}[!t]
\centering
\footnotesize
\resizebox{\textwidth}{!}{
\begin{tabular}{l cccccccc}
\toprule
\textbf{Dataset} & \textbf{Raon} & \textbf{Qwen2.5} & \textbf{Audio} & \textbf{Step-Audio} & \textbf{Interactive} & \textbf{Fun-Audio} & \textbf{HyperCLOVA} & \textbf{MiniCPM} \\
 & \textbf{-Speech} & \textbf{-Omni} & \textbf{Flamingo3} & \textbf{2 Mini} & \textbf{Omni} & \textbf{Chat} & \textbf{X 8B Omni} & \textbf{-o 4.5} \\
\midrule
\multicolumn{9}{c}{\textit{English SpokenQA (Performance $\uparrow$)}} \\
\midrule
OpenBookQA  & \secondbest{86.2} & 83.3 & 62.4 & 77.1 & 76.5 & 81.8 & 29.0 & \textbf{87.7} \\
MMSU  & \textbf{67.5} & 56.3 & 48.6 & 54.5 & 58.7 & 65.3 & 30.1 & \secondbest{66.7} \\
BBH  & \textbf{85.1} & 52.9 & 29.2 & 45.9 & 39.7 & \secondbest{67.7} & 37.9 & 55.0 \\
LlamaQ  & \textbf{81.3} & 77.0 & 69.0 & 72.3 & 78.0 & \secondbest{80.7} & 76.7 & 80.7 \\
TriviaQA  & 62.0 & 58.5 & 28.5 & 52.3 & 57.1 & \secondbest{67.3} & 44.3 & \textbf{68.7} \\
WebQ  & 60.1 & 59.0 & 21.8 & 52.4 & 54.6 & \secondbest{62.2} & 45.6 & \textbf{65.9} \\
SD-QA  & 60.9 & 54.1 & 38.2 & 36.4 & 43.6 & \secondbest{61.7} & 42.7 & \textbf{68.4} \\
AlpacaEval  & 77.4 & 72.4 & 36.2 & 60.6 & 77.0 & \secondbest{79.4} & 63.2 & \textbf{84.0} \\
CommonEval  & 69.2 & 68.0 & 50.4 & 32.4 & \secondbest{70.0} & 68.6 & 62.4 & \textbf{71.8} \\
WildVoice  & \secondbest{70.0} & 62.6 & 42.6 & 41.8 & 63.2 & 66.6 & 51.8 & \textbf{71.6} \\
IFEval  & \secondbest{78.1} & 51.4 & 18.1 & 47.5 & 45.1 & 76.6 & 35.4 & \textbf{80.6} \\
AdvBench  & 96.7 & \textbf{99.4} & 48.7 & 56.2 & 87.9 & 95.2 & 85.8 & \secondbest{98.9} \\
\cmidrule{1-9}
VoiceBench  & \textbf{76.8} & 66.7 & 41.6 & 50.3 & 62.4 & 73.6 & 48.7 & \secondbest{76.1} \\
OpenAudioBench  & 70.2 & 66.7 & 38.9 & 59.6 & 66.7 & \secondbest{72.4} & 57.4 & \textbf{74.8} \\
\midrule
\multicolumn{9}{c}{\textit{Korean SpokenQA (Performance $\uparrow$)}} \\
\midrule
KOpenBookQA  & \textbf{74.8} & \secondbest{31.0} & 6.5 & 7.4 & 7.4 & 28.3 & 27.4 & 25.2 \\
KMMSU  & \textbf{55.5} & 28.3 & 10.0 & 11.7 & 7.0 & \secondbest{28.7} & 24.0 & 26.9 \\
KBBH  & \textbf{83.5} & 49.9 & 36.3 & 40.3 & 8.3 & 52.3 & 52.7 & \secondbest{59.1} \\
KLlamaQ  & \textbf{62.7} & 47.9 & 11.3 & 35.9 & 9.5 & 51.8 & \secondbest{60.2} & 38.4 \\
KTriviaQA  & \textbf{35.5} & 18.0 & 4.5 & 17.4 & 3.6 & 23.7 & \secondbest{26.9} & 22.7 \\
KWebQ  & \textbf{45.3} & 35.1 & 5.3 & 27.3 & 5.9 & 35.4 & \secondbest{38.5} & 30.0 \\
KSD-QA  & \textbf{44.8} & 28.0 & 7.5 & 23.3 & 5.6 & \secondbest{32.1} & 29.1 & 24.8 \\
KAlpacaEval  & \textbf{65.0} & 56.0 & 29.4 & 43.4 & 26.8 & \secondbest{61.4} & 54.8 & 51.6 \\
KCommonEval  & \secondbest{57.4} & \textbf{58.8} & 27.2 & 32.4 & 22.8 & 56.8 & 54.2 & 43.6 \\
KWildVoice  & \textbf{59.6} & 50.6 & 26.0 & 35.2 & 23.0 & \secondbest{53.4} & 48.0 & 41.0 \\
KIFEval  & \textbf{71.6} & 42.9 & 20.8 & 30.4 & 10.7 & \secondbest{53.4} & 32.6 & 34.6 \\
KAdvBench  & \secondbest{87.3} & \textbf{95.9} & 5.7 & 64.3 & 68.0 & 84.8 & 83.2 & 48.5 \\
\cmidrule{1-9}
KVoiceBench  & \textbf{66.6} & 49.0 & 18.8 & 32.0 & 20.0 & \secondbest{50.1} & 45.1 & 39.5 \\
KOpenAudioBench  & \textbf{52.1} & 39.2 & 12.6 & 31.0 & 11.5 & 43.1 & \secondbest{45.1} & 35.7 \\
\bottomrule
\end{tabular}
}
\caption{Detailed SpokenQA results on English source benchmarks and Korean transferred benchmarks. \textbf{Bold} and \underline{underline} indicate the best and second-best per row. Accuracy (\%) is used for multiple-choice and short-answer tasks; GPT-5.4 judge scores (100-point scale) are used for open-ended tasks; prompt/instruction-level accuracy average is used for IFEval/KIFEval; refusal rate is used for AdvBench/KAdvBench.}
\label{tab:detailed_spokenqa_results}
\end{table*}
\begin{table*}[!t]
\centering
\footnotesize
\resizebox{\textwidth}{!}{
\begin{tabular}{ll cccccccc}
\toprule
& & \textbf{Raon} & \textbf{Qwen2.5} & \textbf{Audio} & \textbf{Step-Audio} & \textbf{Interactive} & \textbf{Fun-Audio} & \textbf{HyperCLOVA} & \textbf{MiniCPM} \\
\textbf{Category} & \textbf{Capability} & \textbf{-Speech} & \textbf{-Omni} & \textbf{Flamingo3} & \textbf{2 Mini} & \textbf{Omni} & \textbf{Chat} & \textbf{X 8B Omni} & \textbf{-o 4.5} \\
\midrule
\multirow{3}{*}{Acoustic} & Age & \textbf{62.0} & 41.3 & 21.0 & \secondbest{52.2} & 29.0 & 47.5 & 21.0 & 51.8 \\
 & Gender & \secondbest{93.7} & \textbf{98.9} & 57.0 & 88.9 & 46.3 & 87.8 & 30.0 & 91.1 \\
 & Number of Speakers & 30.0 & 29.0 & 26.0 & 28.0 & 23.0 & \secondbest{33.0} & 21.0 & \textbf{36.0} \\
\midrule
\multirow{6}{*}{Contextual} & Word Order & \textbf{98.7} & 80.4 & 54.1 & \secondbest{94.0} & 52.6 & 74.1 & 15.0 & 54.3 \\
 & Word Frequency & \textbf{42.8} & 34.4 & 13.9 & 25.6 & 21.1 & \secondbest{42.2} & 18.9 & 26.1 \\
 & Fact Extraction & 86.9 & 81.8 & 59.6 & 79.8 & 33.3 & \textbf{91.9} & 36.4 & \secondbest{88.9} \\
 & Topic Summary & 92.0 & 82.0 & 72.0 & 91.0 & 32.0 & \secondbest{96.0} & 69.0 & \textbf{98.0} \\
 & Role/Profession & 75.0 & 80.0 & 69.0 & \secondbest{83.0} & 14.0 & \textbf{92.0} & 44.0 & 78.0 \\
 & General Counting & \secondbest{65.4} & 38.5 & 34.6 & 40.4 & 25.0 & \textbf{69.2} & 28.8 & 42.3 \\
\midrule
\multicolumn{2}{l}{\textbf{Average}} & \textbf{71.8} & 62.9 & 45.2 & 64.8 & 30.7 & \secondbest{70.4} & 31.6 & 62.9 \\
\bottomrule
\end{tabular}
}
\caption{Detailed KMMAU results (accuracy \%) by capability. \textbf{Bold} and \underline{underline} indicate the best and second-best per row.}
\label{tab:kmmau}
\end{table*}
\section{Detailed Evaluation Results}
\label{sec:appendix_eval}

Tables~\ref{tab:detailed_spokenqa_results} and~\ref{tab:kmmau} provide the sub-benchmark and capability-level results omitted from the main table.
The SpokenQA table makes the English-to-Korean shifts visible at the source-task level, while the KMMAU table separates acoustic and contextual audio-understanding capabilities.

\end{document}